# Frequency-Domain Membership Inference Does Not Beat Confidence Thresholding: A Band-Resolved Analysis of Why

Mohammad Zare
Department of Computer Engineering and Information Technology
Shiraz University of Technology,
Shiraz, Iran
md.zare@sutech.ac.ir

Pirooz Shamsinejadbabaki
Department of Computer Engineering and Information Technology
Shiraz University of Technology,
Shiraz, Iran
p.shamsinejad@sutech.ac.ir

**A widely repeated account of membership inference holds that models lean on fine-grained detail for samples they have memorized, so degrading an input's resolution should expose membership through how quickly a prediction falls apart. We test that account band by band. A monotone family of zero-phase low-pass filters is applied to each query image, the target's confidence on its original prediction is read at every step, and the resulting decay is scored in a dozen ways. Two federated ResNet-18 targets are studied, one trained normally and one pushed to a 45.1-point generalization gap. The account does not survive. On the normally trained target the best profile score reaches an AUC of 0.513 against 0.537 for a single confidence threshold, and on the overfit target 0.735 against 0.943. A search over 1206 candidate scores, with direction chosen on a held-out half and a fitted combination as an upper bound, produces no score that beats the threshold in either regime. The failure has a mechanism: the mean confidence gap between members and non-members grows as detail is removed, exactly as the hypothesis predicts, but the spread grows faster, so discriminability falls from the first step onward. Decomposing each image into phase and magnitude shows why no band-localized operation can work, since the advantage travels with image structure, surviving at 0.567 when the phase is kept and collapsing to 0.492 when only the magnitude is. Membership is a whole-image identity effect rather than an accumulation of spectral energy in any band, and defenses aimed at high-frequency reliance are aimed at the wrong target.**

## I. Introduction

A model behaves differently toward a record it was trained on than toward one it has never seen, and membership inference turns the difference into a decision [1], [2]. In federated learning the question has a practical edge: the global model is a function of every participating client's data, and an adversary who can query it inherits much of the exposure a centrally trained model would have had [3], [4]. Work since has mostly weakened the assumptions the adversary needs, from shadow models [1], [5] to single-query statistics [2], [6] to label-only access [7], [8], [9]. Much less has gone into asking which property of the model the signal is attached to. How unevenly training fits its records is a property of the learning procedure rather than of the data alone [10], [11].

A specific answer has circulated. Networks fit low frequencies before high ones and acquire fine detail late, in the regime where test error has stopped improving [12], [13], and the features a network uses divide into a robust class that survives perturbation and a non-robust class that does not [14], [15]. Together these suggest that a memorized sample leans on fine detail more heavily than an unseen one, so removing detail should hurt a member more. The consequence for attack design would be immediate: membership readable from how a prediction decays along a scale ladder, and a single confidence threshold, which sees only the starting point of that decay, leaving something on the table. A resolution-based attack along these lines has been put forward for federated targets [16], which makes the underlying question worth isolating rather than assuming.

This paper tests the proposition and reports that it is false. The test gives it every advantage. Membership is measured band by band through a monotone family of zero-phase low-pass filters, so no band can be missed for want of a suitable operator. The same forward passes that produce the band profile also produce the confidence and entropy thresholds, so the comparison against prior training-free attacks uses identical samples rather than a matched draw. Two targets are studied, one trained normally and one deliberately overfit until its generalization gap reaches 45 percentage points, because a memorization effect should be easiest to see where memorization is extreme. Direction and any fitted weights come from a held-out half of the samples, and every reported number comes from the other half.

The profile does not win. On the normally trained target it lands below the confidence threshold, and on the overfit target, where a single threshold already reaches an AUC near 0.94, the best profile score manages 0.74. A linear combination fitted over the full feature set does worse still, which rules out the possibility that the right weighting was simply not found. The failure has a clean mechanism and the mechanism is the more interesting result. The mean confidence gap between members and non-members does grow as detail is removed, exactly as the hypothesis predicts, but the spread grows faster, so discriminability falls from the first step onward. Chasing that further produces a second finding: decomposing each image into phase and magnitude shows the advantage travels with image structure and vanishes when only the energy is kept, so no band-localized operation can isolate it.

## II. Related Work

Membership inference began with shadow models, which train auxiliary networks on data similar to the target's and fit a classifier on their outputs [1], later refined by reducing the shadow budget or replacing the classifier with a likelihood ratio across many reference models [5], [17]. The cost is real: the strongest such formulation trains hundreds of models per target, hard to justify against a federated target that was itself expensive to produce.

A second line removes auxiliary training. A single-query threshold on loss or confidence separates members from non-members well enough to be useful [2], threshold attacks survive dataset shift better than shadow classifiers [6], and output entropy does about as well as confidence [6], [18]. These are the right baselines for any new training-free method, and their weakness is structural: a scalar response cannot express how a prediction was reached, so two samples with equal confidence are indistinguishable no matter how differently the model arrived there. Closing that gap is what a multi-step profile is supposed to do, and it is what this paper tests. Label-only attacks probe instead of reading, counting label flips under augmentation [7], explanations [8], or a single crafted query [9]; their perturbations are usually random or adversarial, and a structured ladder over spatial scale is the alternative examined here.

In federated settings passive and active membership inference both remain effective when only updates are exchanged [3], updates leak more than membership [4], and gradients can be inverted outright [19], while poisoning the training set can be turned into a membership oracle [20]. Coarser client-level inference has been studied too [21], and the field's open problems have been catalogued [22]. Defenses have followed: differential privacy bounds any single record's influence [23], [24], distillation reduces memorization [25], and MemGuard perturbs outputs to break the confidence signal [26], with surveys covering the area [27]. Federated learning has also been applied to image generation with privacy preservation as an explicit design goal [28]. A recurring assumption across this literature, and the one put under test here, is that residual exposure is mediated by a model's reliance on sample-specific detail, with the corollary that detail-level operations are the right place to look.

The spectral side of that assumption has support from two directions. Networks fit low frequencies first, acquiring high-frequency components once test error has stopped falling [12], [13], and predictive features split into a robust class that survives perturbation and a non-robust class that does not [14], [15]. Frequency-domain tools have long been used to decompose images along these lines: scale-space representations [29], Laplacian pyramids [30], and the classical result that phase carries an image's structure while magnitude carries its contrast envelope [31]. Colourisation methods depend on the same asymmetry [32]. Separately, the aliasing of strided convolution has been treated as a robustness defect, corrected by a fixed low-pass [33], [34]. What has not been done is the measurement. Degradation-based signatures have been proposed for this setting [16], and the question they raise, whether the decay along a scale ladder carries anything the first response does not, has not been answered band by band against a threshold computed from the same forward passes.

*Table 1. Attack families and the question this study adds.*

| Attack family | Aux. training | Queries | Signal used | Band-resolved |
|---|---|---|---|---|
| **Shadow models [1], [5]** | Hundreds of models | 1 | Attack classifier | No |
| **One-shot statistics [2], [6]** | None | 1 | Loss, confidence, entropy | No |
| **Label-only [7], [8], [9]** | None | Many | Label flips | No |
| **Active federated [3]** | None | Per round | Update inspection | No |
| **This study** | None | $K + 1$ | Band-resolved decay, tested | Yes |

## III. Measurement Design

### A. Setting and Notation

Let $f$ be the target classifier, mapping an image to a probability vector over $C$ classes, queryable on inputs of the adversary's choosing. We write $x0$ for the query image, $y^* = \text{argmax}_y\, f_y(x0)$ for the class predicted on the unmodified input, and $\rho$ for radial frequency in cycles per image, so that $\rho = 0$ is DC and $\rho = H/2$ is the Nyquist limit along an axis. The adversary scores images with no shadow model, no auxiliary data and no gradient. The question under study is whether anything measured along a scale ladder carries information that the confidence $f_{y^*}(x0)$ available at the first query does not.

### B. Scale Ladder

Scale ladders in this literature usually downsample and restore, which has a defect once the ladder is long enough to matter: repeated dyadic downsampling from 32×32 reaches 1×1 after five steps, at which point the image is a single colour and the measurement is a property of the dataset's colour statistics rather than of the sample. We use a frequency-domain low-pass with a continuous cutoff instead. With $X$ the centered discrete Fourier transform of the image and $\rho(u,v)$ the distance of bin $(u,v)$ from DC, the response $H(\rho; r, n) = 1 / (1 + (\rho/r)^{2n})$ attenuates frequencies above the cutoff radius $r$ with a roll-off set by $n$, and the eroded image is the inverse transform of the product, clipped to $[0,1]$. The response is real and even, so the filter has zero phase and cannot shift the image, which matters because a phase-shifting operator would confound attenuation with translation. The ladder is $r_k = r_{max} (r_{min}/r_{max})^{(k-1)/(K_{max}-1)}$, with $r_{max}$ at the Nyquist radius and $r_{min} = 1$. Every experiment that varies the number of steps uses a prefix of that one sequence, so $K$ measures how many steps the measurement takes rather than a change of spacing.

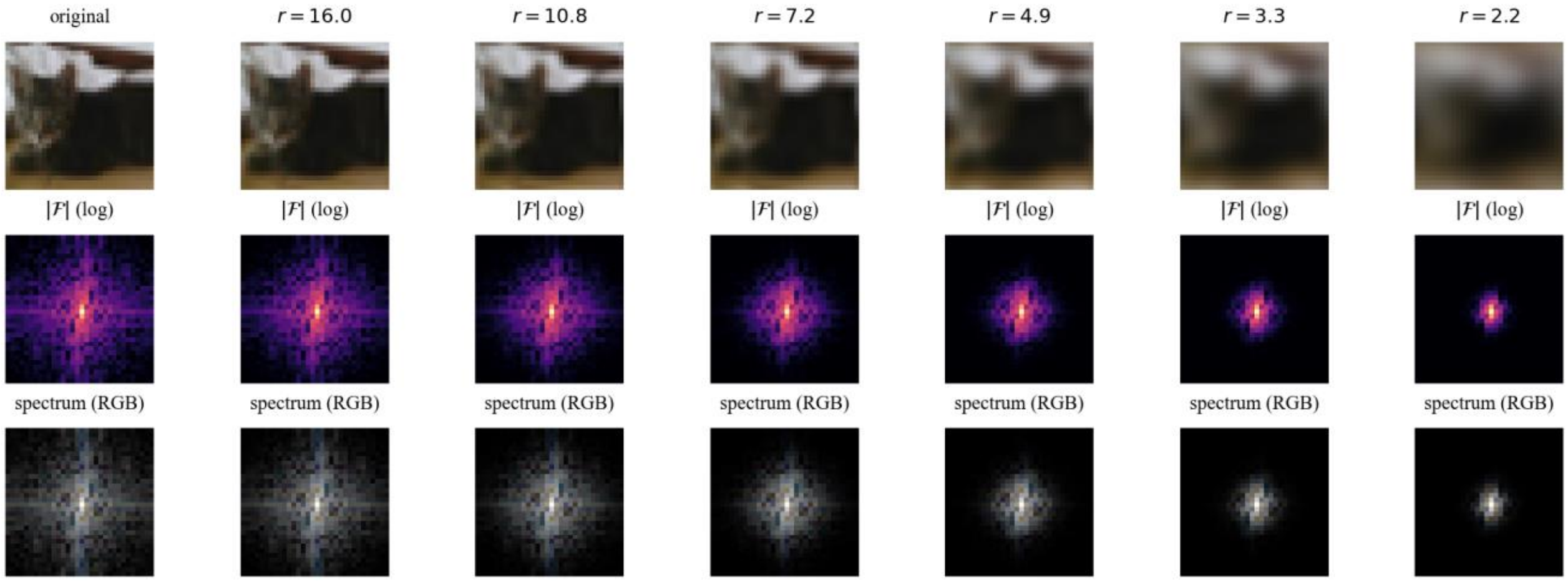

*Fig. 1. A CIFAR-10 image and its log-magnitude spectrum at successive erosion steps. The spectrum contracts toward DC as the cutoff radius falls.*

### C. Wavelet and Spatial Controls

A dyadic wavelet decomposition orders an image by scale by another route. The orthonormal Haar transform splits an image into a coarse approximation and three detail subbands per level, level one holding the finest detail, and zeroing the L finest detail levels while reconstructing preserves a comparable range of scales through an exact invertible decomposition rather than a filtered one. We include it so the result cannot be attributed to a particular filter family, and include spatial average pooling as well, since that is the operator the claims under test were originally made with.

### D. Profile and Scores

At each step the model is queried and we read the probability it assigns to $y^*$, the class it predicted for the untouched image. Writing $c_k = f_{y^*}(x_k)$ for the confidence after k steps, the decay is $s_k = c_{k-1} - c_k$ for $k = 1, \ldots, K$ and the profile is the sequence of these. Fixing the class is deliberate: recomputing the argmax after each step would mix the decay of a decision with changes of the decision. The spectra leakage area summarizes the profile as a scalar, in its simplest form the mean per-step loss $SLA = (1/K)\, \Sigma_k\, s_k$. Because that reduction discards the profile's shape, which is the only thing the multi-step measurement adds, Section IV also tests the components $s_k$ individually, a weighted form with $w_k$ proportional to $1/k$, a length-normalized form $(c_0 - c_K)/(K c_0)$, and the profile's Euclidean norm.

### E. Phase and Magnitude Decomposition

The second half of the study asks what the advantage is attached to, and for that we use the decomposition of Oppenheim and Lim [35]. Given a sample x and a fixed real reference r, a phase-kept reconstruction takes the sample's phase with the reference's magnitude, and a magnitude-kept reconstruction takes the sample's magnitude with the reference's phase. The phase-kept image remains recognisable to a human observer while the magnitude-kept one does not, which is what makes the pair a clean probe. Because r is fixed across samples, any difference between member and non-member responses within a condition is attributable to the sample. The reference choice matters: swapping against the mean training image, whose magnitude is concentrated at low frequencies, turns the phase-kept condition into a low-pass and confounds the test with the ladder experiment, so the primary reference is a real photograph and the mean-image swap is reported separately as the low-pass limit.

### F. Threat Model and Protocol

The adversary observes softmax confidences, the assumption made by the one-shot and shadow-model attacks we compare against [1], [2], [6] and available through most prediction APIs. Membership is at the item level, the adversary queries the final global model, and aggregation-level defenses are outside scope. Every score costs $K + 1$ forward passes and no training. Because this is a measurement rather than a deployed attack the protocol matters more than the threat model: scores are compared on identical samples, the baselines come from the same forward passes as the profile, and a paired test is used rather than overlapping intervals.

### G. How the Search Is Bounded

A negative result is only as strong as the search behind it. Samples are split into a development half and a reporting half by a fixed permutation. The profile is read at every step for the full probability vector, from which a library is derived: confidence on $y^*$, maximum probability, entropy, the top-1 minus top-2 margin, the rank of $y^*$, total variation from the first step's distribution, Kullback-Leibler divergence, cosine similarity, and the number of steps before the predicted class changes. Each is taken raw, differenced against step zero, and divided by step zero, then summarized by its first value, its first two, its mean, its last, and a fixed fine-weighted average, crossed with every ladder length. Direction is chosen on the development half so a score informative in the reverse orientation is credited, and the reported AUC is always from the other half. Two guards are added: every candidate is also evaluated after the sign is fixed, on the full sample, with DeLong's paired test against the confidence threshold, and a logistic regression fitted on the development half over the entire feature matrix is reported as a ceiling that a training-free attacker could not reach.

### H. Computational Cost

The profile costs $K + 1$ forward passes per sample, one for the untouched image and one per step. The filtering runs in the Fourier domain on a $32 \times 32$ image and costs a fraction of a forward pass. Nothing is trained and the queries are independent across samples.

# IV. Results

## A. Setup

Two federated targets are trained on CIFAR-10 [35], which holds 50,000 training and 10,000 test images of 32×32 pixels across ten classes. The training split is partitioned uniformly across ten clients and a ResNet-18 [36] is trained with FedAvg [37], one local epoch per round, with standard crop and flip augmentation. At 120 rounds the global model reaches 91.7% test and 97.2% training accuracy, a gap of 5.4 points. Three seeds are run with the client partition fixed. That target sits in the regime where membership inference is known to be hard, which on its own would make a negative result uninformative, so a second target is built to make memorization extreme: the same architecture and procedure on 5,000 samples rather than 50,000, augmentation disabled. It reaches 100.0% training and 54.9% test accuracy, a gap of 45.1 points. Membership is evaluated on 1000 samples drawn from the training split, spread evenly across clients, against 1000 from the held-out test split, and the same pools serve every score and ablation so comparisons are paired.

## B. The Profile Does Not Beat the Threshold

Table 2 reports the comparison. On the normally trained target the confidence threshold reaches an AUC of 0.537 and entropy 0.537, while the best of the seven profile scores reaches 0.513, a loss that is not small relative to the seed-to-seed spread of 0.003. On the overfit target the ordering holds with more room: the threshold reaches 0.943, close to perfect on this pool, while the best profile score manages 0.735. At the operating points that matter for a privacy audit the picture is the same: holding the false positive rate at 1% recovers 0.9% of members from the threshold and 1.0% from the profile on the standard target, and 2.1% against 6.5% on the overfit one. DeLong's test confirms the differences are not sampling noise, and the direction is against the profile in every seed of both regimes.

One nuance belongs here because it would be easy to overstate. On the overfit target the profile's true positive rate at a 1% false positive rate is 6.5% against 2.1% for the threshold, even though its AUC is far lower. The profile has heavier tails, identifying a small additional set of members with unusual trajectories at the cost of separating the bulk much worse. With 1,000 non-members a 1% rate rests on ten samples, so this is coarsely estimated and is not a reason to prefer the profile as an attack. It is a reason to note that a low-false-positive audit and an average-case benchmark are not the same measurement.

*Table 2. The band profile against the single-query thresholds, both regimes. Mean over seeds.*

| Target | Score | AUC | TPR@FPR=1% (%) |
|---|---|---|---|
| **Standard (gap 5.4)** | Confidence threshold | 0.537 ± 0.005 | 0.9 |
| | Entropy threshold | 0.537 ± 0.005 | 0.7 |
| | Best band profile | 0.513 ± 0.003 | 1.0 |
| **Overfit (gap 45.2)** | Confidence threshold | 0.943 ± 0.000 | 2.1 |
| | Entropy threshold | 0.942 ± 0.000 | 2.0 |
| | Best band profile | 0.735 ± 0.000 | 6.5 |

## C. Why: the Gap Grows While Discriminability Falls

The failure is not a matter of the score being badly chosen, and the reason it cannot be fixed by better scoring is visible in the trajectory itself. Two quantities are easy to conflate: the raw gap between the mean confidence of members and non-members at a given step, and that gap measured in units of the pooled spread, which is what governs whether the two distributions can be separated.

On the overfit target the raw gap behaves exactly as the hypothesis predicts, rising from 0.2159 at the first step to 0.2935 at step 3, an increase of 36%, which is to say that removing fine detail does hurt members more in absolute terms. Over the same steps the normalized gap falls from 1.467 to 1.081 and the AUC of the confidence level falls from 0.943 to 0.789. The standard target shows the same pattern in miniature, its raw gap peaking at step 2 while the AUC is already below its starting value. Erosion does not shift members and non-members apart so much as it reshuffles both. Blurring a member's image removes the evidence the model was responding to; blurring a non-member's removes whatever weak evidence it had. The member mean falls less over the mild steps, which is why the raw gap grows, but the variance of both grows faster, because after enough erosion the residual confidence is driven by colour statistics that vary far more across images than membership does. A larger gap inside a much wider distribution is a worse measurement. This also disposes of a natural repair: per-sample normalization does not recover the signal, since dividing the total decay by the initial confidence leaves an AUC of 0.500 on the standard target, indistinguishable from chance, and the profile norm does no better at 0.516.

*Table 3. The member and non-member confidence gap, the same gap in units of the pooled spread, and the AUC of the confidence level.*

| Step k | Raw gap (std) | Gap/spread (std) | AUC (std) | Raw gap (over) | Gap/spread (over) | AUC (over) |
|---|---|---|---|---|---|---|
| **0** | 0.0168 | 0.149 | 0.537 | 0.2159 | 1.467 | 0.943 |
| **1** | 0.0211 | 0.138 | 0.535 | 0.2257 | 1.450 | 0.940 |
| **2** | 0.0269 | 0.094 | 0.522 | 0.2648 | 1.396 | 0.913 |
| **3** | 0.0127 | 0.031 | 0.509 | 0.2935 | 1.081 | 0.789 |
| **4** | 0.0036 | 0.011 | 0.507 | 0.1644 | 0.433 | 0.625 |
| **6** | 0.0064 | 0.026 | 0.503 | 0.0130 | 0.035 | 0.504 |
| **8** | 0.0060 | 0.028 | 0.504 | -0.0064 | -0.019 | 0.490 |

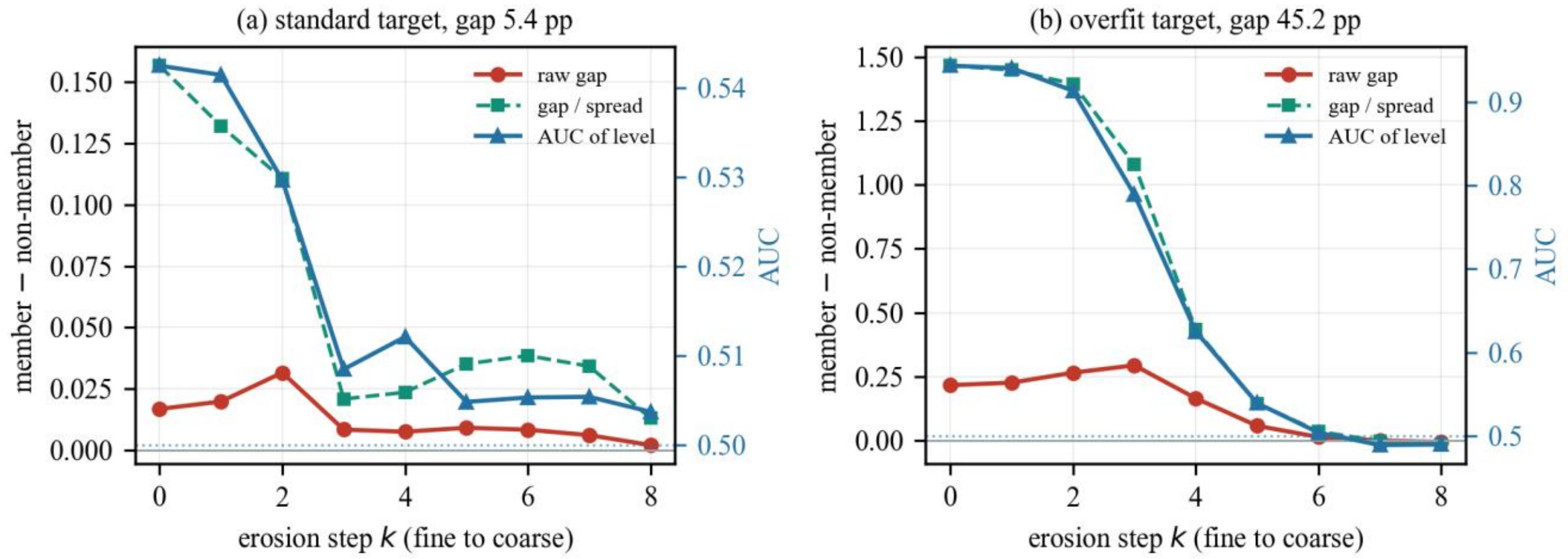

*Fig. 2. The raw member and non-member confidence gap against the discriminability of the same quantity, along the ladder. The raw gap rises while the AUC falls.*

### D. Band Attribution and the Reversal

Reading the profile one band at a time runs against the hypothesis rather than merely failing to support it. On the overfit target, where the signal is strong enough for individual bands to be informative, the three finest bands have AUCs of 0.457, 0.413, 0.464, all below chance, while the mid bands reach 0.628. A band whose AUC is below chance is not an uninformative band but a band whose decay is systematically larger for non-members, so removing the finest detail costs a non-member more confidence than it costs a member, the opposite of what a fine-detail memorization account requires. The standard target shows the same ordering compressed against the noise floor. Two caveats: adjacent bands are not independent, since a smooth filter attenuates neighbouring frequencies together, so the per-band AUCs locate the signal rather than decompose it, and the ordering is a statement about this ladder at this resolution.

*Fig. 3. AUC from a single band's decay. On the overfit target the three finest bands fall below chance, so their decay is larger for non-members than for members.*

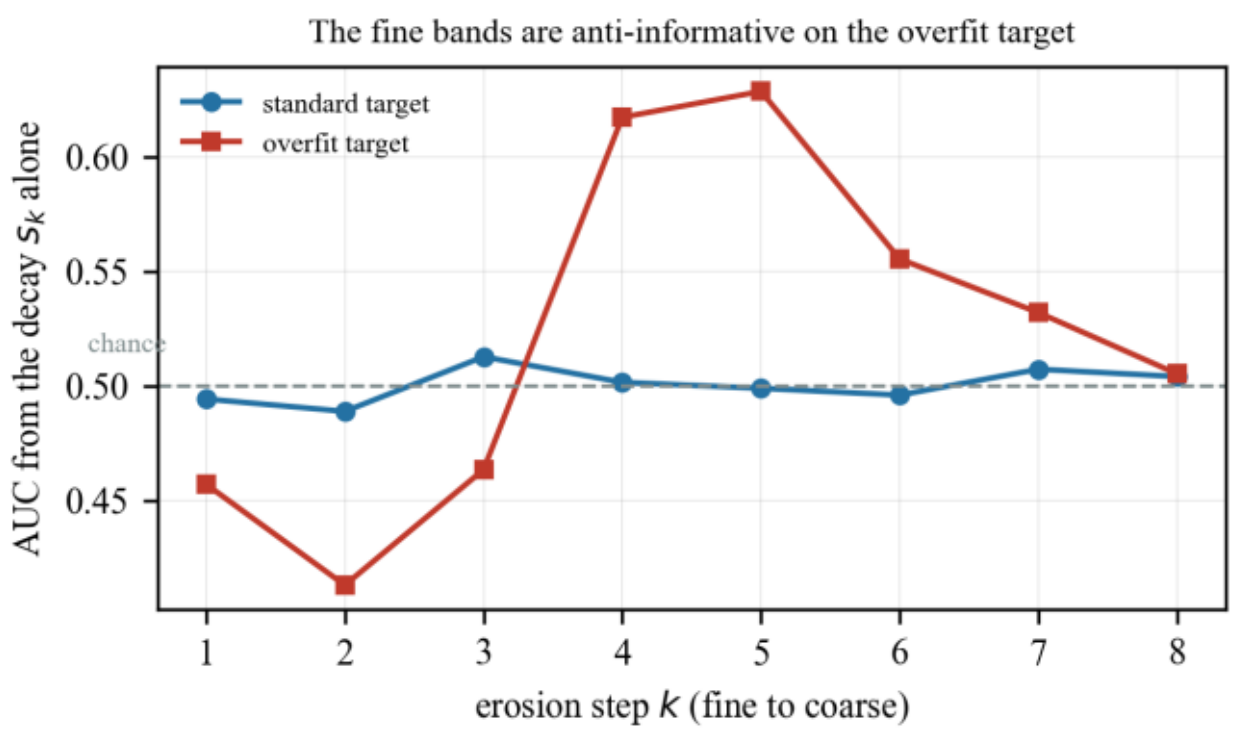

### E. Phase and Magnitude

The band result raises the question of what the advantage is attached to if not to spectral energy. On the overfit target the original images give 0.943, the phase-kept reconstructions give 0.567, and the magnitude-kept reconstructions give 0.492. Keeping the structure and discarding the energy preserves a substantial part of the signal; keeping the energy and discarding the structure destroys it entirely. On the standard target the same contrast appears in weaker form (0.506 against 0.502), and the mean member and non-member confidences tell the same story. The size of the phase-kept figure is as important as its direction. Recovering 0.57 against an original 0.94 means that preserving structure while replacing the contrast envelope already costs most of the signal. The advantage is not a band, not a summary statistic of the spectrum, and not even the sample's structure in isolation; it is the model's response to this particular image with these particular values. That is why a ladder over spatial scale cannot isolate it: there is no band whose removal separates members from non-members, because the property being measured is not distributed across bands to begin with.

*Table 4. AUC under the phase and magnitude swap against a fixed real reference photograph.*

| Target | Reconstruction | AUC |
|---|---|---|
| **Standard** | Original image | 0.537 ± 0.005 |
| | Phase kept | 0.506 ± 0.005 |
| | Magnitude kept | 0.502 ± 0.007 |
| **Overfit** | Original image | 0.943 ± 0.000 |
| | Phase kept | 0.567 ± 0.000 |
| | Magnitude kept | 0.492 ± 0.000 |

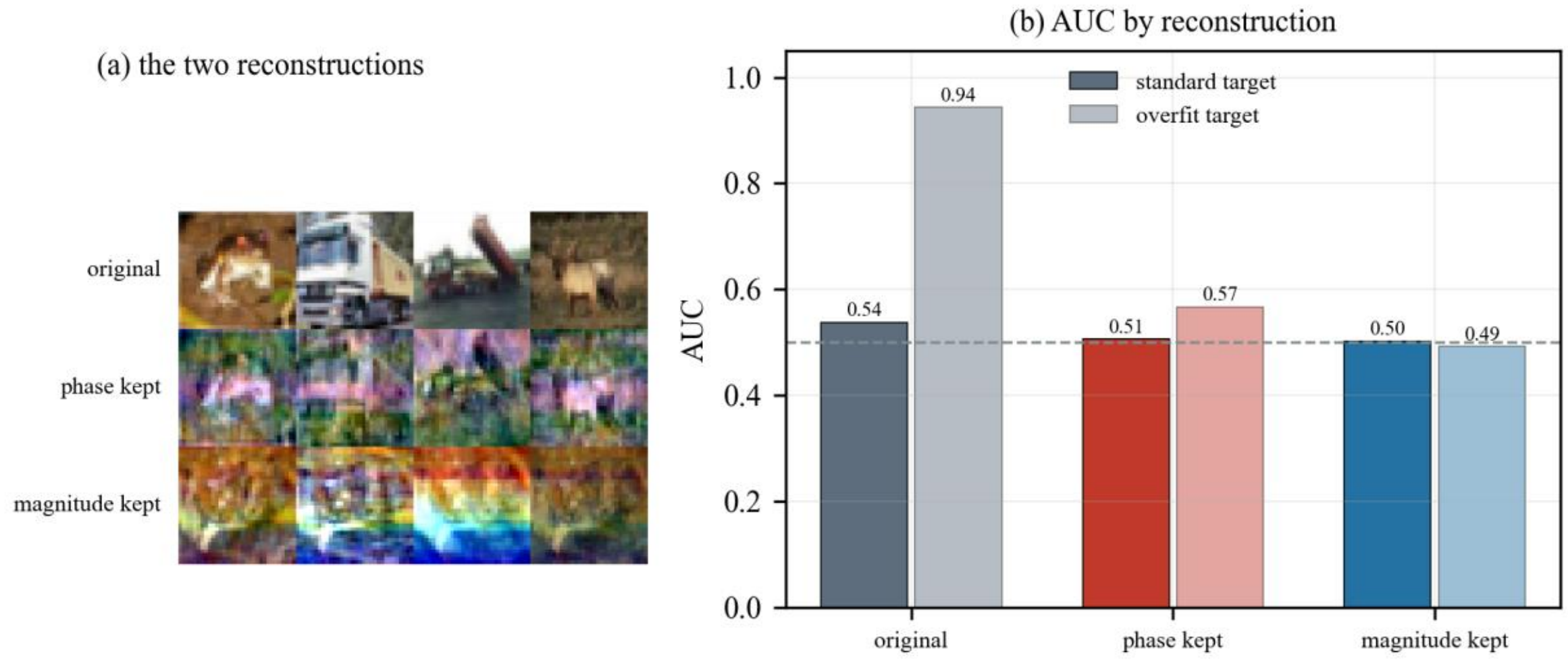

*Fig. 4. (a) The two reconstructions. The phase-kept row keeps the structure of the original. (b) AUC under each reconstruction, for both targets.*

### F. The Search and Its Ceiling

The search evaluates 1206 candidates in each regime: every derived quantity from the full probability vector at every step, in raw, differenced and ratio form, summarized five ways over steps, crossed with every ladder length. Two results settle the question. First, the best candidate found in the standard regime reaches 0.518 on the development half where it was selected against 0.563 on the reporting half, a shrinkage that is what selection noise looks like; the winner differs per seed and the apparent gain does not survive when the same score is evaluated on all samples with DeLong's test. Second, the fitted ceiling. A logistic regression over the entire stacked feature matrix, given labels the attacker would not have, reaches 0.499 on the standard target, worse than chance, and 0.937 on the overfit target, below the 0.943 that a single confidence value achieves alone. A model with access to twelve hundred measurements per sample and to the membership labels themselves cannot extract more than the first coordinate already contains. The standard-target figure signals that the fit overfits the development half and generalizes poorly, which is why the fitted model is reported as a ceiling that happened to be unreachable rather than as evidence of anything; the overfit-regime figure carries the argument, since there the fit has a strong signal to find and still does not exceed the threshold.

*Table 5. The bounded search. Selection uses the development half; the reported columns are from the untouched half.*

| Target | Candidates | Best on dev | Same score on test | Fitted ceiling |
|---|---|---|---|---|
| **Standard** | 1206 | 0.518 | 0.563 | 0.499 |
| **Overfit** | 1206 | 0.950 | 0.941 | 0.937 |

### G. Ablations

Three ablations establish that the result belongs to the setting rather than to the implementation. Holding the scale range fixed and changing how detail is removed moves the AUC from 0.521 for Spatial avg-pool to 0.513 for Butterworth, and every operator lands below the confidence threshold, including the spatial average pooling the claims under test were originally made with. Extending the ladder from one step to 8 moves the AUC from 0.498 to 0.513, never recovering the threshold's performance and flattening after the third step. Varying the Butterworth order from one to four moves it from 0.514 to 0.513, so the result is not an artifact of an unlucky roll-off. Client behaviour is uniform as well: scoring each client's members against the pooled non-members gives AUCs spanning 0.467 to 0.555, narrow enough that no client stands out, which matters because a federated negative result would otherwise be open to a client-identity reading.

*Table 6. Ablations. (a) ladder length, (b) erosion operator, (c) roll-off, (d) client spread.*

| Setting | Value | AUC |
|---|---|---|
| **K = 1** | 1 step | 0.498 |
| **K = 3** | 3 steps | 0.504 |
| **K = 8** | 8 steps | 0.513 |
| **Spatial avg-pool** | 8 steps | 0.521 |
| **Butterworth** | 8 steps | 0.513 |
| **Order n = 1** | roll-off | 0.514 |
| **Order n = 4** | roll-off | 0.513 |
| **Clients** | 10 per-client AUCs | 0.467 to 0.555 |

## V. Discussion

The confidence threshold is not a weak baseline that a cleverer statistic will eventually displace. On these targets it is close to the information the query budget can supply, and the fitted ceiling suggests that is a property of the signal rather than of our score library. An attack that reads only the first response and stops has not left anything obvious on the table. What it has left, on this evidence, is a small set of samples with unusual trajectories that show up at very low false positive rates, which is a narrower opportunity than a general improvement and is worth pursuing as such.

If membership leakage is an image-identity effect rather than an accumulation of high-frequency energy, then interventions that remove or suppress high-frequency content are addressing a channel that carries little of the signal. Anti-aliased downsampling, which inserts a fixed low-pass before every stride and is known to improve shift invariance [33], [34], is the natural candidate for a frequency-targeted defense, and this study gives no reason to expect it to help. The interventions that should work are the ones acting on how strongly the model can fit any single record, which is what differential privacy does [23] and what the generalization gap measures. The two targets here differ by a factor of eight in that gap and the achievable attack AUC differs by a factor of nearly two, with no need to look at the spectrum to predict either.

Two habits in this literature would have caught this earlier. The first is reporting a degradation-based attack without computing the single-query threshold from the same forward pass, which makes the comparison exact. The second is reporting the mean separation between members and non-members without the spread, since in this setting the two move in opposite directions and the mean alone tells the wrong story. Neither habit is specific to frequency-domain attacks. Everything reported here comes from CIFAR-10 at 32×32 pixels, a ResNet-18 and an approximately uniform client partition. Higher-resolution data carries more high-frequency content and could in principle behave differently, although the mechanism identified here does not depend on resolution. Non-IID partitions, where client distributions differ [38], and architectures whose inductive bias differs from a convolutional network are untested, transformers in particular processing spatial information without the translation structure that makes a scale ladder natural. We also tested one augmentation setting for the standard target, and augmentation strength is known to change how much a model memorizes, which is exactly the axis along which our two regimes differ.

## VI. Conclusion

We asked whether membership information is distributed across spatial scale in a way a ladder of low-pass filters could read, and found that it is not. On a federated ResNet-18 trained on CIFAR-10, no band profile, no summary of one, and no fitted combination of twelve hundred derived quantities beats the confidence the model reports on its first response, which reaches 0.537 on the standard target and 0.943 on a target trained until its generalization gap is 45.1 points.

The mechanism is the part worth keeping. The mean confidence gap between members and non-members does grow as fine detail is removed, which is what the fine-detail account predicts, but the spread grows faster, so the gap becomes less usable even as it becomes larger. And the advantage itself travels with image structure rather than with spectral energy, surviving at 0.567 when the phase of a sample is preserved against 0.492 when only its magnitude is. Membership, in the models we measured, is not a property of any part of the spectrum but of the whole image, and it is lost the moment the image stops being the one the model saw. That reframes what a frequency-domain defense can hope to achieve, and it suggests that the most informative next measurement is the same decomposition applied to a target whose privacy budget is actually bounded, where the question becomes how much of the identity effect differential privacy removes rather than whether the effect exists.